\documentclass[twocolumn]{article}
\pdfoutput=1
\usepackage[utf8]{inputenc}
\usepackage[T1]{fontenc}
\usepackage{lmodern}                
\usepackage[english]{babel}
\usepackage{hyperref}           
\usepackage{parskip}            
\usepackage{microtype}          
\usepackage{graphicx}
\usepackage{abstract}
\usepackage{subfig}
\usepackage{float}
\usepackage[backend=bibtex,natbib=true,sorting=none]{biblatex}
\usepackage{booktabs}
\usepackage[inline]{enumitem}

\newcommand{\tamfig}{0.9}
\newcommand{\tammeiafig}{0.45}
\newcommand{\mycite}{\cite}

\title{Study and development of a Computer-Aided Diagnosis system for classification of chest x-ray images using convolutional neural networks pre-trained for ImageNet and data augmentation}

\author{Vinicius Pavanelli Vianna\thanks{PhD Student at Department of Physics, FFCLRP, University of São Paulo - Ribeirão Preto/SP}\\
\texttt{vnvianna@usp.br}}
\begin{document}
\twocolumn[
    \maketitle
    \begin{onecolabstract}
    \vspace{1em}
Convolutional neural networks (ConvNets) are the actual standard for image recognizement and classification. On the present work we develop a Computer Aided-Diagnosis (CAD) system using ConvNets to classify a x-rays chest images dataset in two groups: \texttt{Normal} and \texttt{Pneumonia}. The study uses ConvNets models available on the \texttt{PyTorch} platform: \texttt{AlexNet}, \texttt{SqueezeNet}, \texttt{ResNet} and \texttt{Inception}. We initially use three training styles: complete from scratch using random initialization, using a pre-trained ImageNet model training only the last layer adapted to our problem (transfer learning) and a pre-trained model modified training all the classifying layers of the model (fine tuning). The last strategy of training used is with data augmentation techniques that avoid over fitting problems on ConvNets yielding the better results on this study.
\vspace{1em}
    \end{onecolabstract}
]
\saythanks
\section{Introduction}
Computer-aided diagnosis (CAD) methods started on the 60's\mycite{CAD_1960} but without great success on that time. Large scale CAD usage arrived in the 80's with the new approach of not replacing the medical professional but only assist in their diagnosis \mycite{CAD_review}. 

Recent growth of computational capacity allowed convolutional neural networks (ConvNets) applications on recognizement and detection of images to expand \mycite{Lecun_2015}, this is true specially after the introduction of \texttt{AlexNet} in 2012 \mycite{ImageNet}. This growth also applies for CAD systems using ConvNets to classify patients.

Normally ConvNets can be subdivided in two networks. One network that extract features from the image, processing attributes like edges in a variety of orientations (e.g. horizontal or vertical ones) to form larger attributes of the images like the presence of square or rounded objects in it. The second network is the classifier that receives the features or attributes processed on the first layer and use all this information to classify the image in classes (i.e. groups of similar images). The process of training the network in extracting the features and classifying is a time-consuming one and demand a large dataset to get good features extracted instead of not so good ones.

To understand the problem of training those networks better the dataset for the ImageNet\mycite{ILSVRC15} competition have 1.000.000 images and even with this large dataset some networks seem to work better than others, even the same network can outperform itself due to a better training. In order to avoid this training problem two important techniques have emerged in reusing ConvNets for classification of small datasets:

\begin{description}
\item[Transfer Learning] is the method of using a network well trained on a similar task by copying its parameters and weights to the adapted network on the current task. Commonly we use the trained network only changing its final classifying layer or layers to solve the current task (e.g. removing the last layer outputting the 1.000 classes on ImageNet problem and adding one layer that outputs just 2 classes for our current problem). We can also handle this method by just adding an extra layer that will receive the original output and convert it to the current problem classes.

\item[Data augmentation] is one of the techniques used on \texttt{AlexNet} \mycite{ImageNet} to improve its training and consist on geometric transformations that generates new training images from the original ones. Examples of those transformation can be doing an horizontal flip of the original images or just cropping and resizing it.

\end{description}

On the current task we will build a CAD system to classify x-rays chest images in two groups: \texttt{Normal} and \texttt{Pneumonia}, using ConvNets and comparing the performance using four different strategies:

\begin{description}
\item[Scratch] Networks initialized from scratch with random parameters, with no prior training.

\item[Transfer Learning] Networks initialized with parameters copied from a network trained for ImageNet, replacing the final layers with new random initialized layers and only training those final layers.

\item[Fine Tuning] Networks also initialized with parameters copied from ImageNet trained ones and with the final layer replaced but now training the classifier or even the entire network.

\item[Data Augmentation] Networks trained with the same strategy as Fine Tuning but now with an augmented dataset, as explained later on.

\end{description}
\section{Methodology}
We used the \texttt{PyTorch}\footnote{\url{https://pytorch.org/}} platform for all our neural network code since it replaced and also uses the code for the \texttt{Caffe2} platform that was used by many other papers in this field.

The dataset used for our study can be found on \mycite{Dataset} and was also used for an study in \mycite{Kermany2018}. Some numbers on this dataset are on Table \ref{tabela_dataset}. On this study we only use the Test and Train sets ignoring the Validation set as it is too small and can't provide a good estimative on the trained network quality or accuracy. The original images have different sizes but all way above the size normally used on ImageNet networks that is $224 \times 224$\footnote{ConvNets normally uses $224 \times 224$ RGB images with exception of Inception that uses $299 \times 299$}.

\begin{table}[ht]
\centering
\begin{tabular}{lrr}
Subset & Original & Balanced\\
\midrule
Validation & 23 & 16\\
Test & 631 & 468 \\
Train & 5,223 & 2,682 \\
\end{tabular}
\caption{Number of chest x-rays images in the dataset from \mycite{Dataset}}
\label{tabela_dataset}
\centering
\begin{tabular}{lrrr}
Network & Top-1 & Top-5 & Reference\\ 
\toprule
\texttt{Inception} v3 & 22.55 & 6.44 & \mycite{inception}\\
\texttt{ResNet} 18 & 30.24 & 10.92 & \mycite{resnet}\\
\texttt{SqueezeNet} 1.1 & 41.81 & 19.38 & \mycite{squeezenet}\\
\texttt{AlexNet} & 43.45 & 20.91 & \mycite{ImageNet}\\
\end{tabular}
\caption{List of the ConvNets used}
\label{tabela_convnets}
\end{table}

We will use the ConvNets listed on Table \ref{tabela_convnets} noting that in the initial phase of our study only some networks were used to minimize time spent on training.\footnote{A complete list of ConvNets available on the platform used can be seen at \url{https://pytorch.org/docs/master/torchvision/models.html}}

The original dataset was unbalanced and for better usage we eliminated exceeding images from the \texttt{Pneumonia} class since it had the greater number. This was a random process made to have an 1:1 ratio between both \texttt{Training} and \texttt{Test} sets.

\begin{figure}[h]
\subfloat[][]{\includegraphics[width=\tamfig\linewidth]{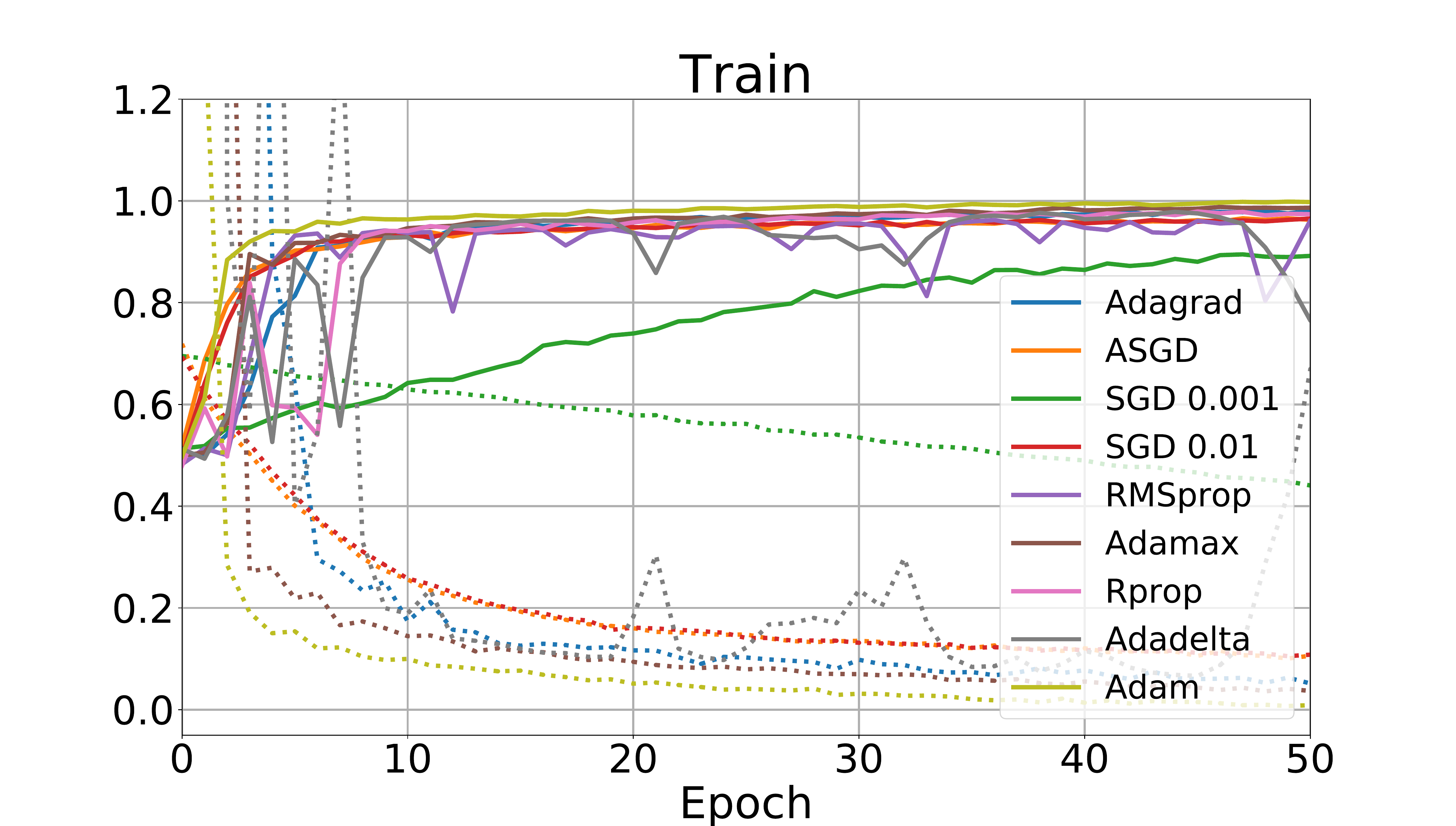}}
\\
\subfloat[][]{\includegraphics[width=\tamfig\linewidth]{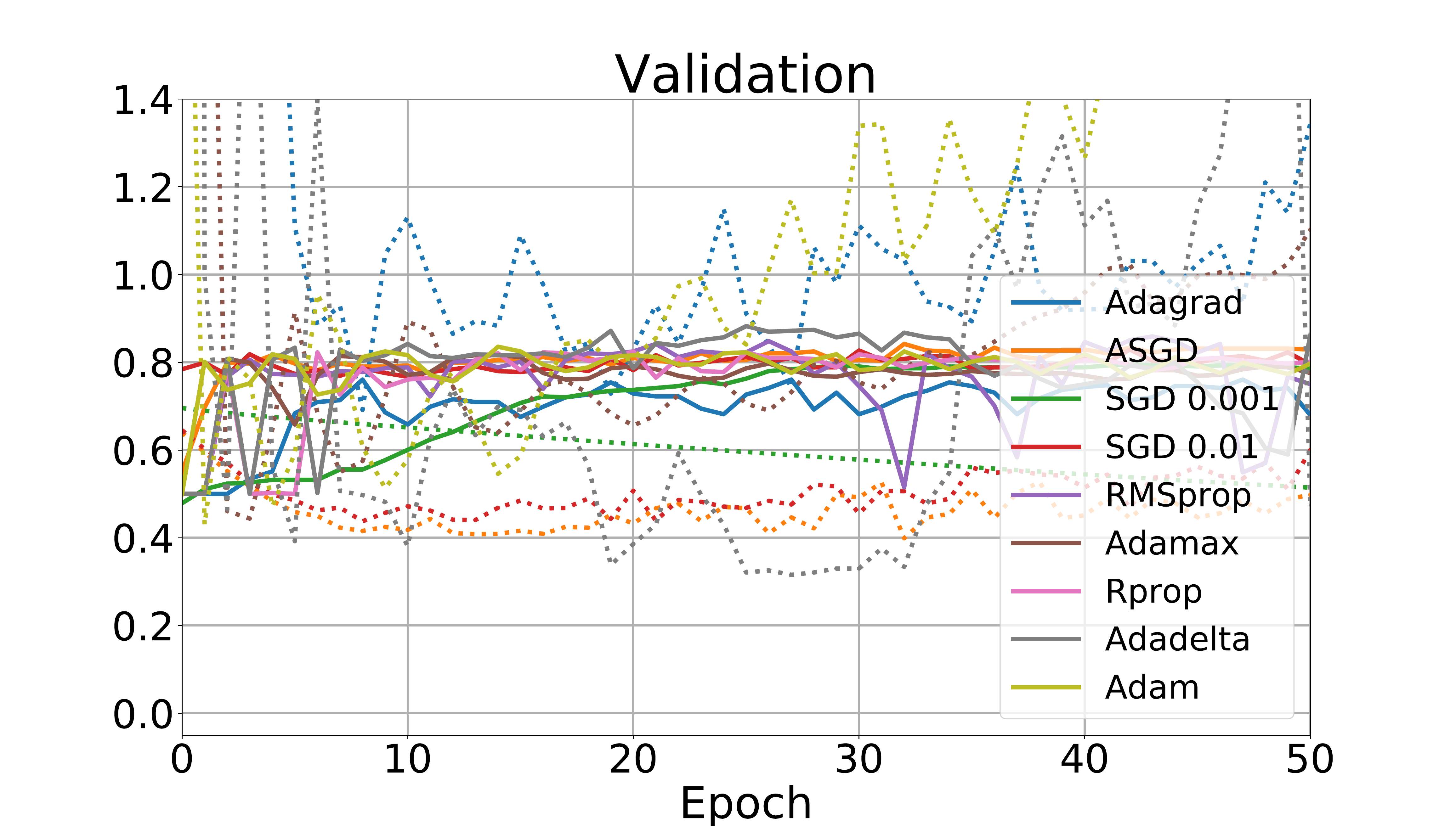}}
\caption{Optimizers comparison results using default parameters except where noted on SGD}
\label{grafico_opt}
\end{figure}

For the optimizer choice to train the network we made some simple tests with all the choices from \texttt{PyTorch} that had a simple and exchangeable operation on our code. The results can be seen on Fig. \ref{grafico_opt}. Overall results are very similar on the long term and with exception of \texttt{SGD} we used the default parameters on the platform. We choose to use the \texttt{Adam} optimizer on the other tests since it showed the fastest convergence with higher accuracy and lower loss on the training set. Here is the full list of optimizers tested:
\begin{enumerate*}[noitemsep,nolistsep,label={\alph*)}]
\item \texttt{ADADELTA} \mycite{adadelta}
\item \texttt{Adagrad} \mycite{adagrad}
\item \texttt{Adam} \mycite{adam}
\item \texttt{Adamax}
\item \texttt{ASGD} \mycite{asgd}
\item \texttt{RMSprop} \mycite{rmsprop}
\item \texttt{Rprop}
\item \texttt{SGD} \mycite{sgd}
\end{enumerate*}

For the transfer learning strategy we made some tests using the many methods seem on other studies, all on the \texttt{AlexNet} model:

\begin{itemize}[noitemsep]
\item Adding an extra layer with 2 neurons connected to the 1.000 outputs of the original last layer, training only this extra layer.

\item Changing the final layer of the original network so it only have 2 neurons getting in this way only 2 output classes, training only this final layer

\item Change the entire \texttt{AlexNet} classifier network, diminishing the number of neurons on the hidden layers to analyse the impact on performance.
\end{itemize}
And for the fine tuning strategy:
\begin{itemize}[noitemsep]
\item Changing the final layer on the original network to only have 2 output neurons but now training the entire classifier network.
\end{itemize}

In the data augmentation strategy we also used the fine tuning strategy by changing the final layer to get 2 output classes but now doing the data augmentation with the following methods from the PyTorch class \texttt{torchvision.transforms}\footnote{\url{https://pytorch.org/docs/master/torchvision/transforms.html}}:
\begin{description}
\item[transforms.RandomResizedCrop()] this method generates a new image by resizing and cropping the original image to the network input image size, using a random resize scale.
\item[transforms.RandomHorizontalFlip()] this method will randomly do the horizontal flip on the image
\end{description}

\section{Results}
\begin{figure}[h]
\subfloat[][]{\includegraphics[width=\tamfig\linewidth]{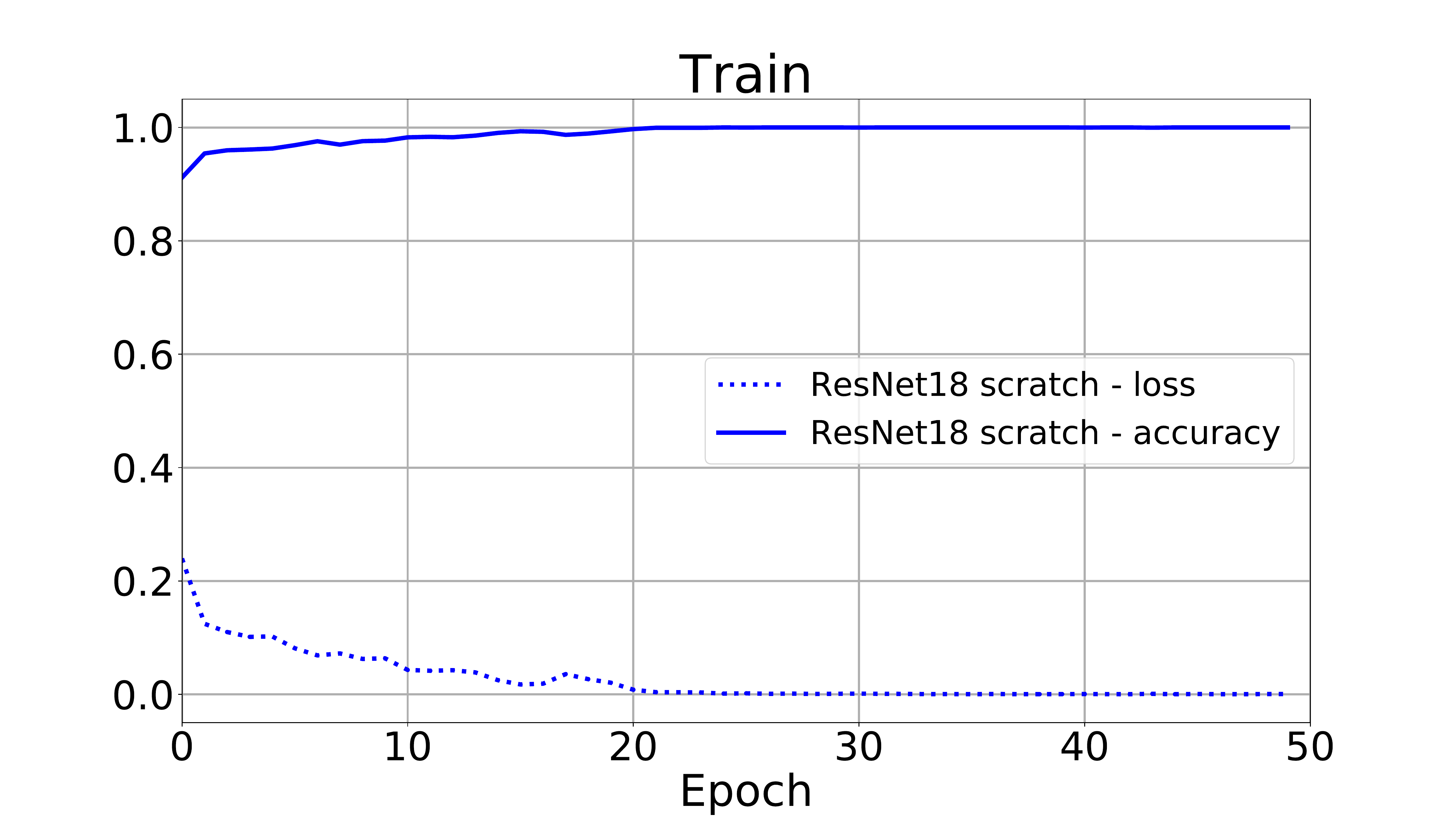}}
\\
\subfloat[][]{\includegraphics[width=\tamfig\linewidth]{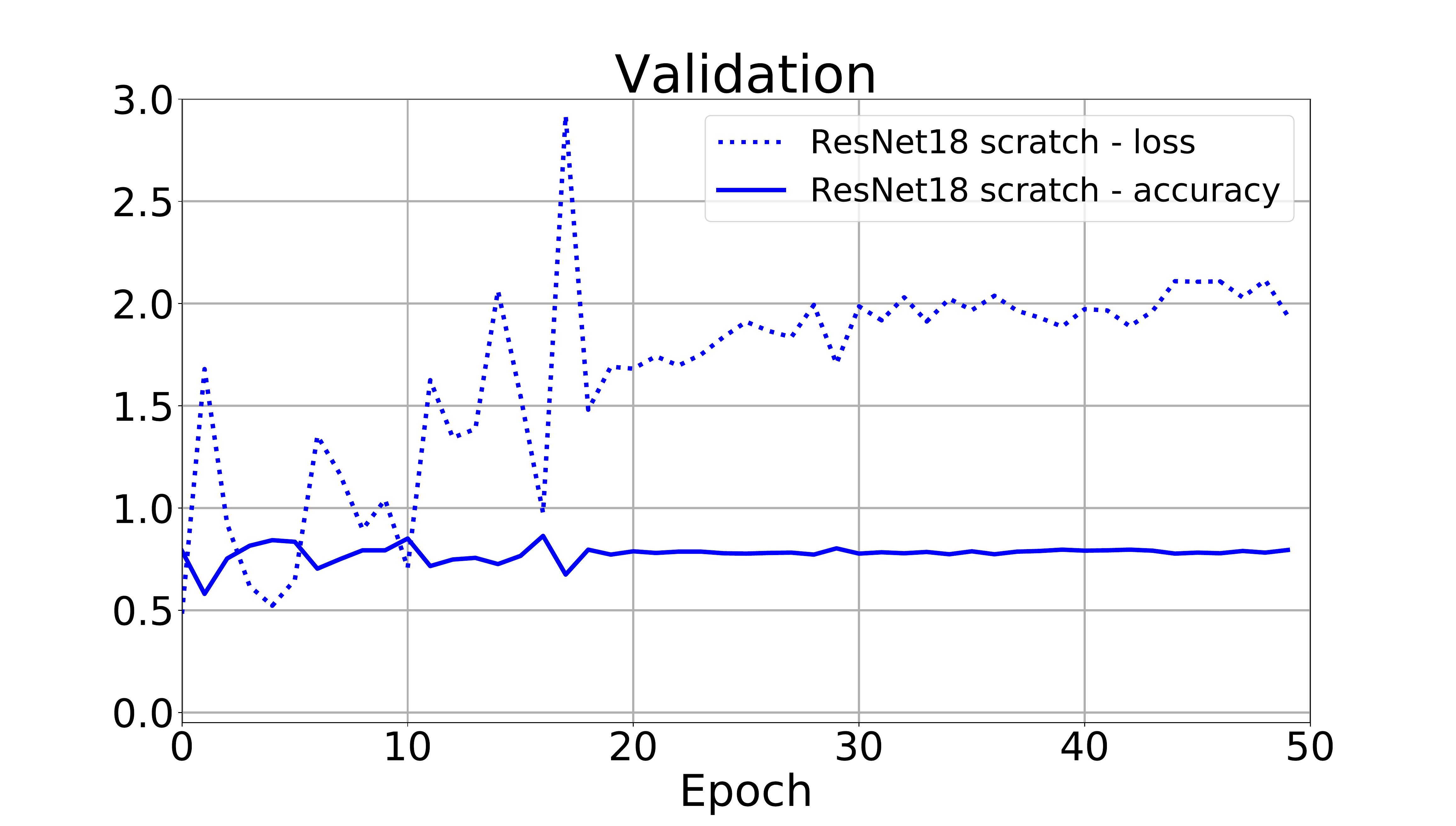}}
\caption{Training results of \texttt{ResNet18} network from scratch}
\label{grafico_zero}
\end{figure}

Using the \texttt{ResNet18} network from scratch we got a maximum validation accuracy of $86.38\%$ (Fig. \ref{grafico_zero}). As the epoch grows the training accuracy got to $100\%$ with its loss getting as low as $0.00034$ but without any reflex on the validation accuracy and a considerable growth in the validation loss. This is a possible evidence of over fitting where the network has adjusted too much to the training set using attributes irrelevant to our classification problem since it didn't help the validation accuracy to increase. Similar results occurred with other networks used in the same strategy.

\begin{figure}[h]
\subfloat[][]{\includegraphics[width=\tamfig\linewidth]{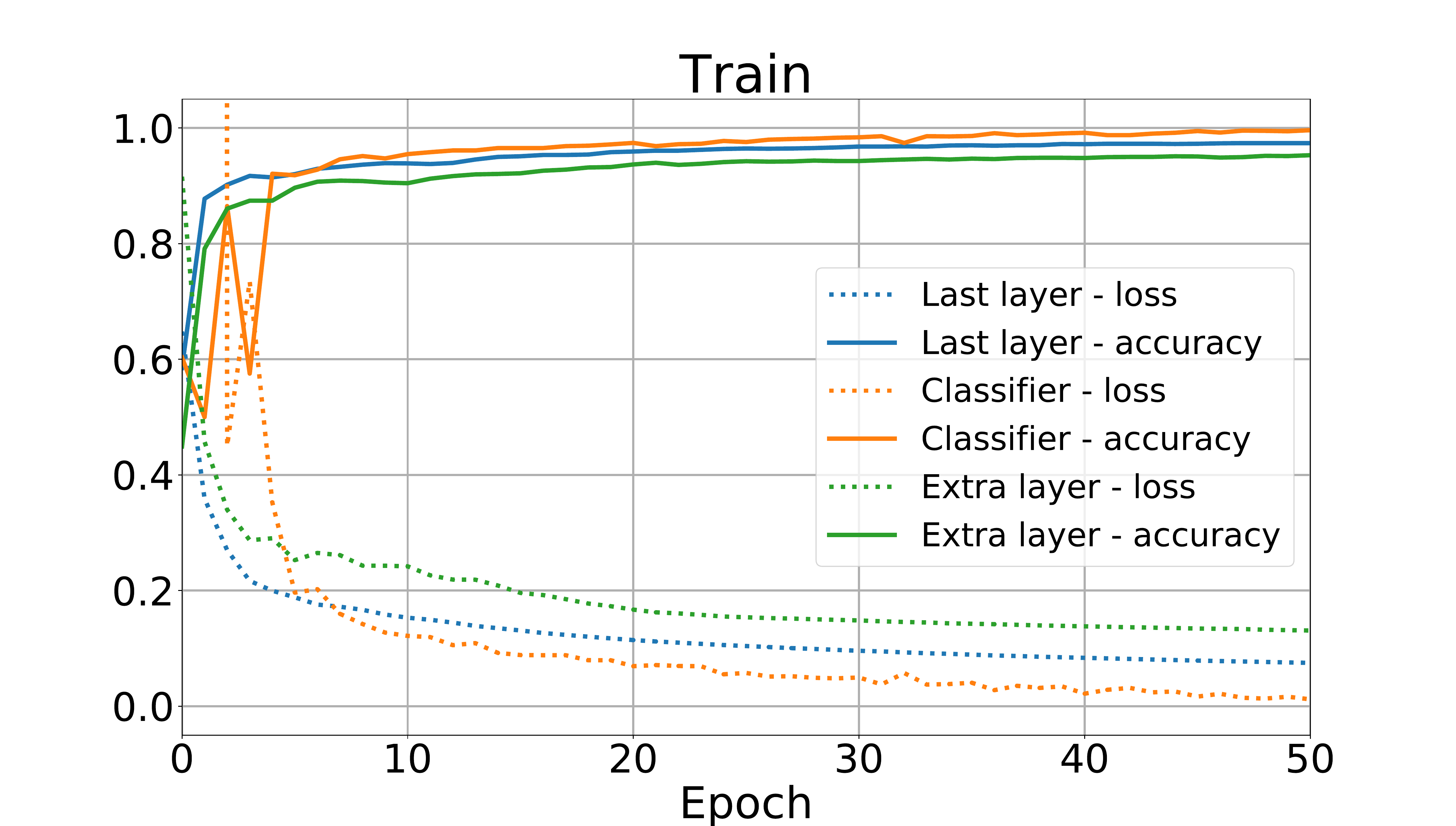}}
\\
\subfloat[][]{\includegraphics[width=\tamfig\linewidth]{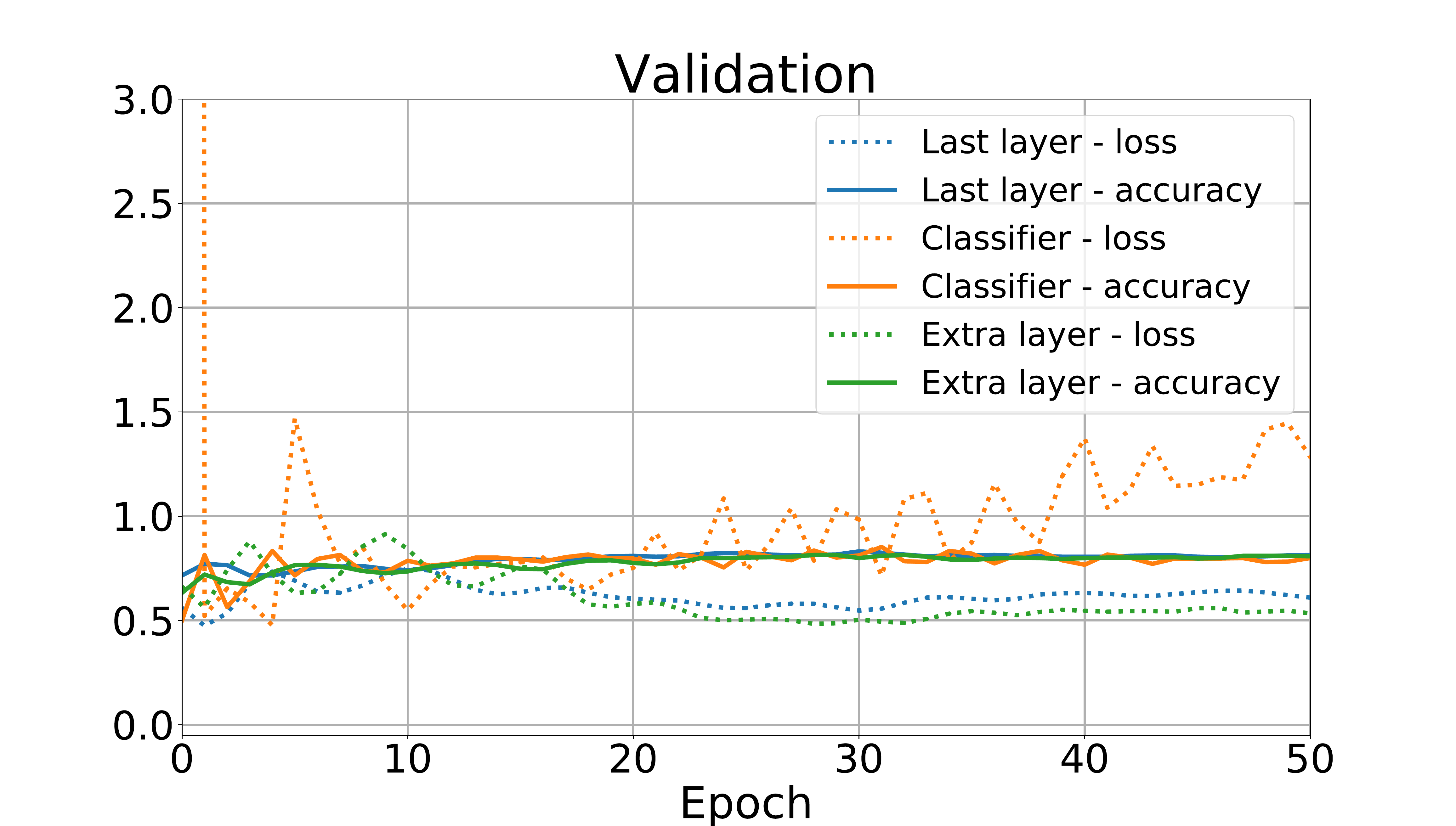}}
\caption{Transfer Learning results using the \texttt{AlexNet} network}
\label{grafico_tl}
\end{figure}

Using the \texttt{AlexNet} network with the transfer learning strategy the accuracy gets a little better than the previous strategy but not all the tests made could reduce the training loss. This shows a limitation on the capacity of a single layer to classify our problem. 

Only with fine tuning when we trained all the classifier network of the \texttt{AlexNet} network we could eliminate the training loss, but in this case we could also see some over fitting on the validation loss. The general validation accuracy of the three approaches to transfer learning where similar as can be seen on Fig. \ref{grafico_tl}.

\begin{figure}[h]
\subfloat[][]{\includegraphics[width=\tamfig\linewidth]{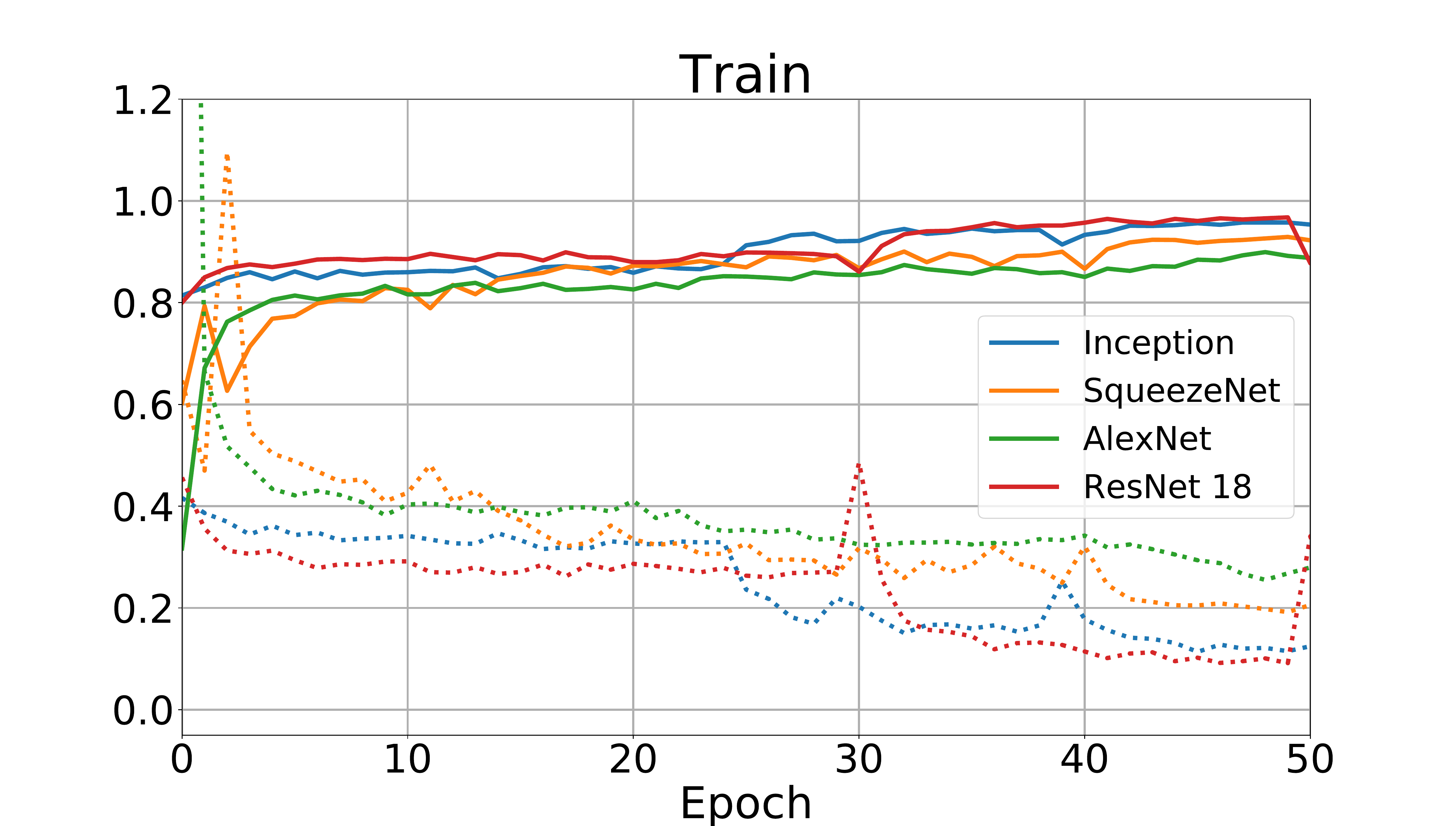}}
\\
\subfloat[][]{\includegraphics[width=\tamfig\linewidth]{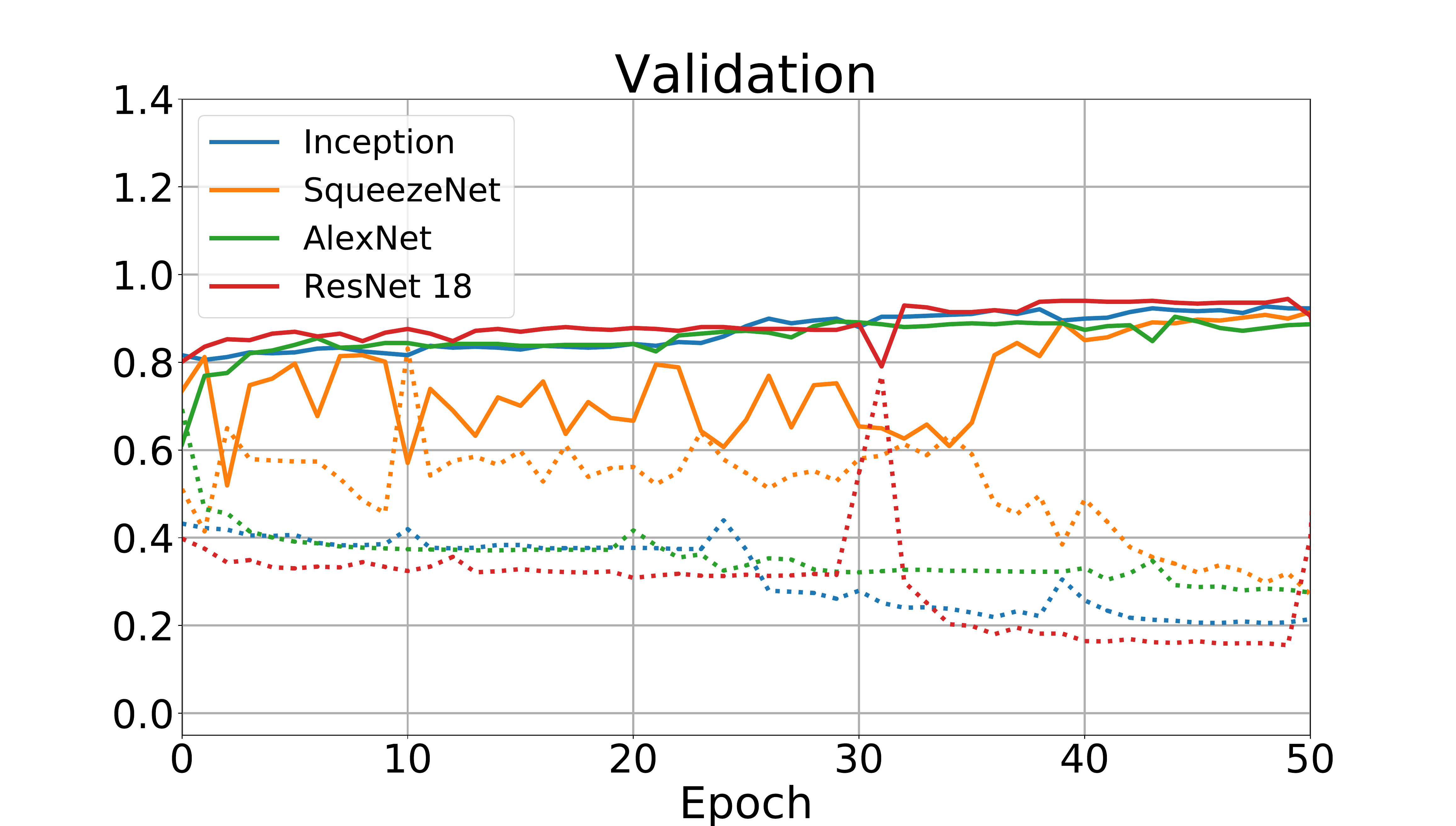}}
\caption{Training results of ConvNets using data augmentation - solid lines are accuracy and dotted lines loss}
\label{grafico_augmentation}
\end{figure}

The best results in accuracy came from the transfer learning technique with data augmentation, as can be seen on Fig. \ref{grafico_augmentation}. The data augmentation technique reduced the over fitting effect on the validation loss and also did not allowed the network to come close to an $100\%$ training accuracy or simply minimized the training loss. This can be credited to the data augmentation not allowing the network to adjust itself to the training set since it's now been changing by data augmentation on every iteration. The best results in terms of accuracy come from the \texttt{ResNet18} ($96.37\%$) and the \texttt{Inception} ($95.51\%$).

\begin{figure}[h]
\includegraphics[width=\tamfig\linewidth]{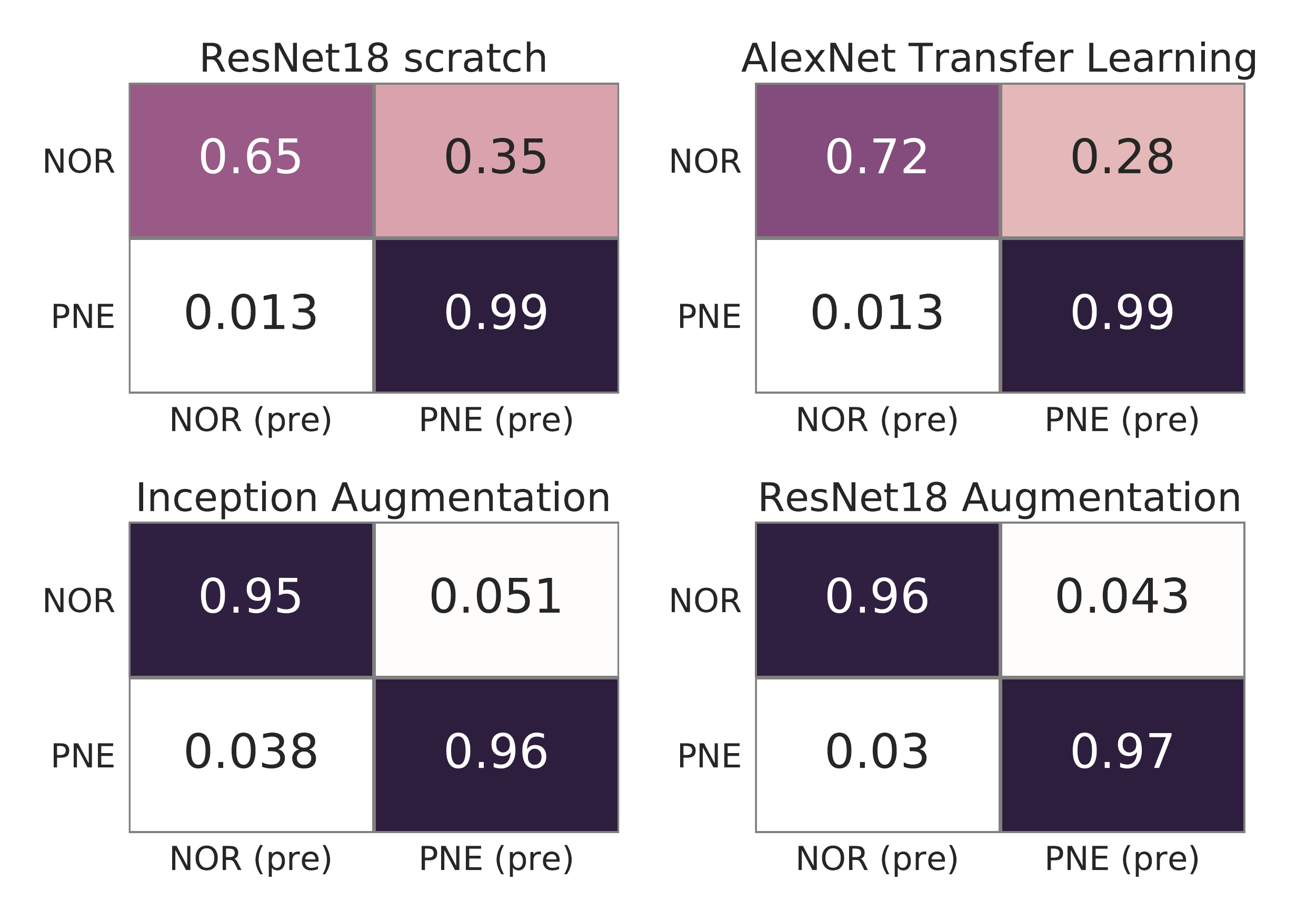}
\caption[Confusion matrices]{Confusion matrices - NOR stands for Normal, PNE for Pneumonia and (pre) means predicted}
\label{grafico_confusao}
\end{figure}

Fig. \ref{grafico_confusao} shows the confusion matrices of different strategies used in this study, we can perceive a high number of false positives in networks without data augmentation (top matrices) even with a high number of hits in the \texttt{Pneumonia} class. When we use the data augmentation technique the false positive and false negative numbers dropped down (bottom matrices) with a high accuracy of both training and validation sets. 

\begin{figure}[h]
\subfloat[][]{\includegraphics[width=\tammeiafig\linewidth]{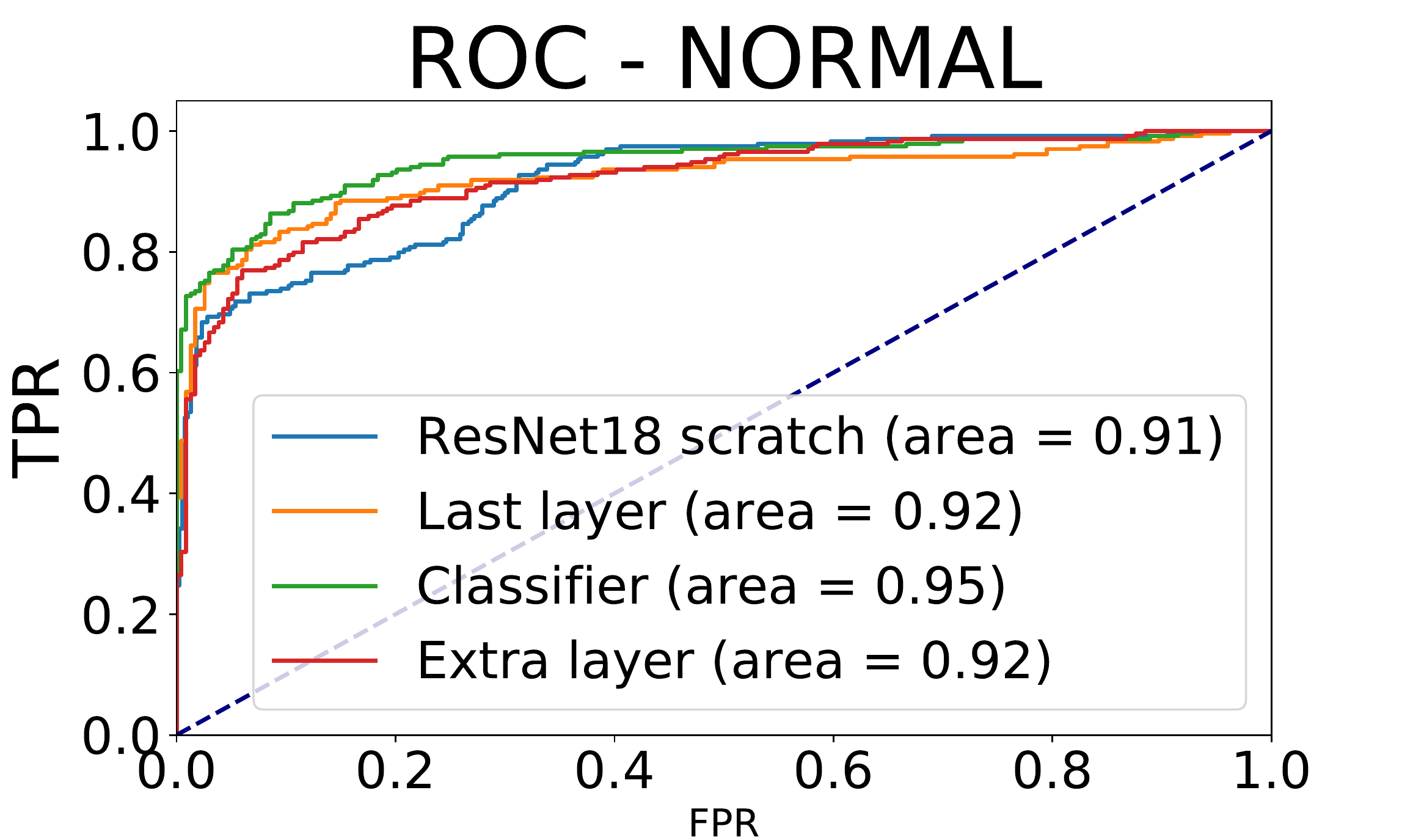}}
\subfloat[][]{\includegraphics[width=\tammeiafig\linewidth]{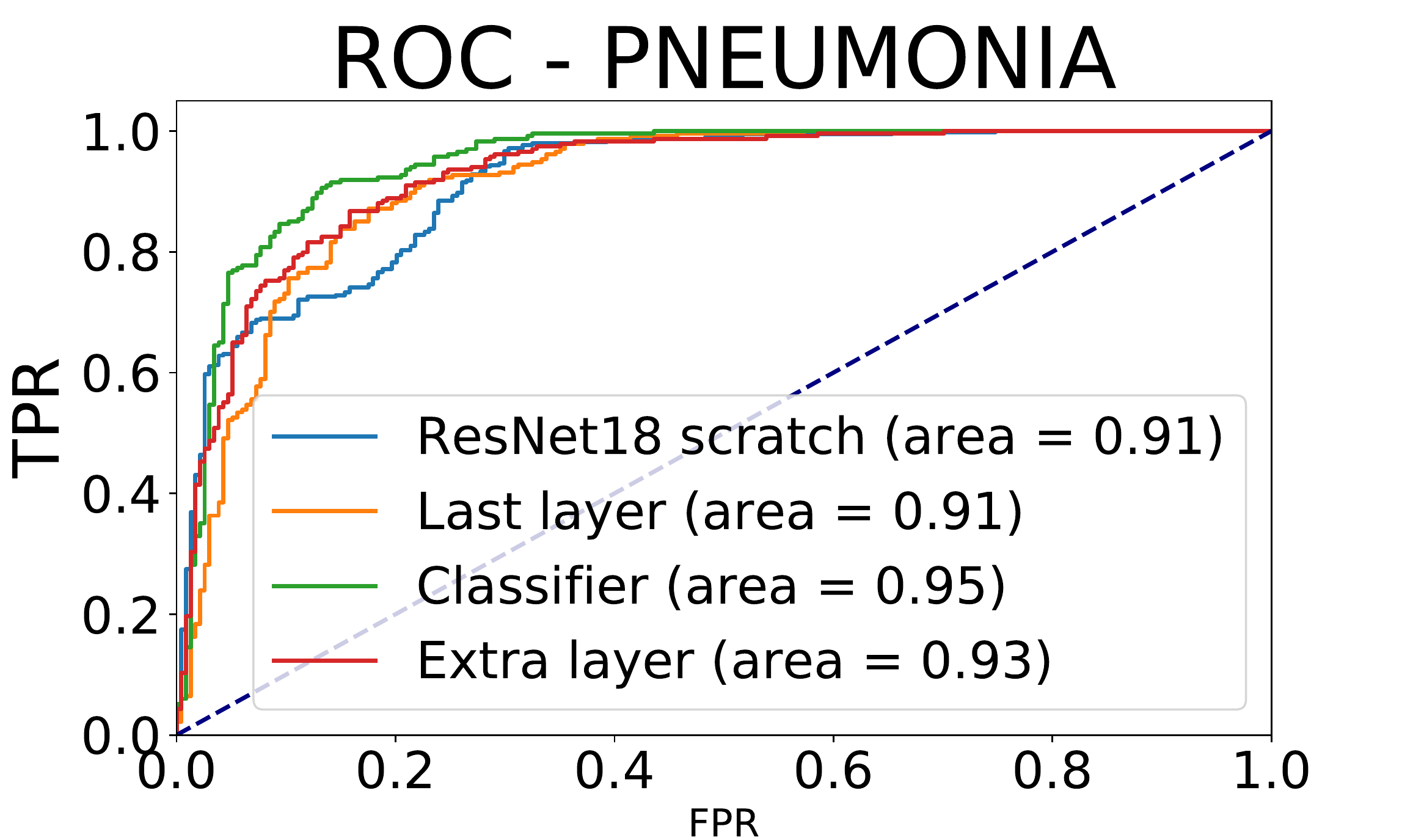}}
\\
\subfloat[][]{\includegraphics[width=\tammeiafig\linewidth]{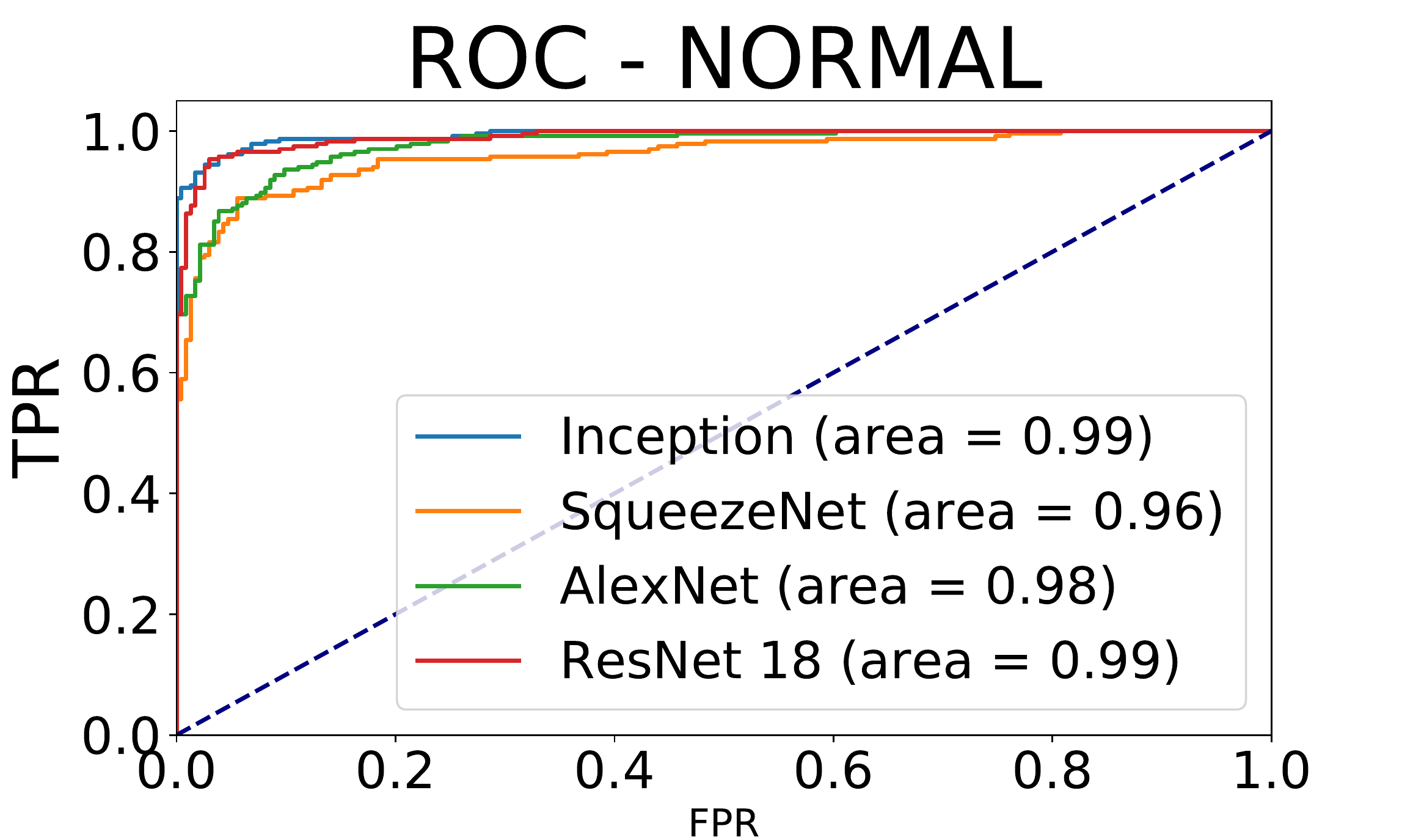}}
\subfloat[][]{\includegraphics[width=\tammeiafig\linewidth]{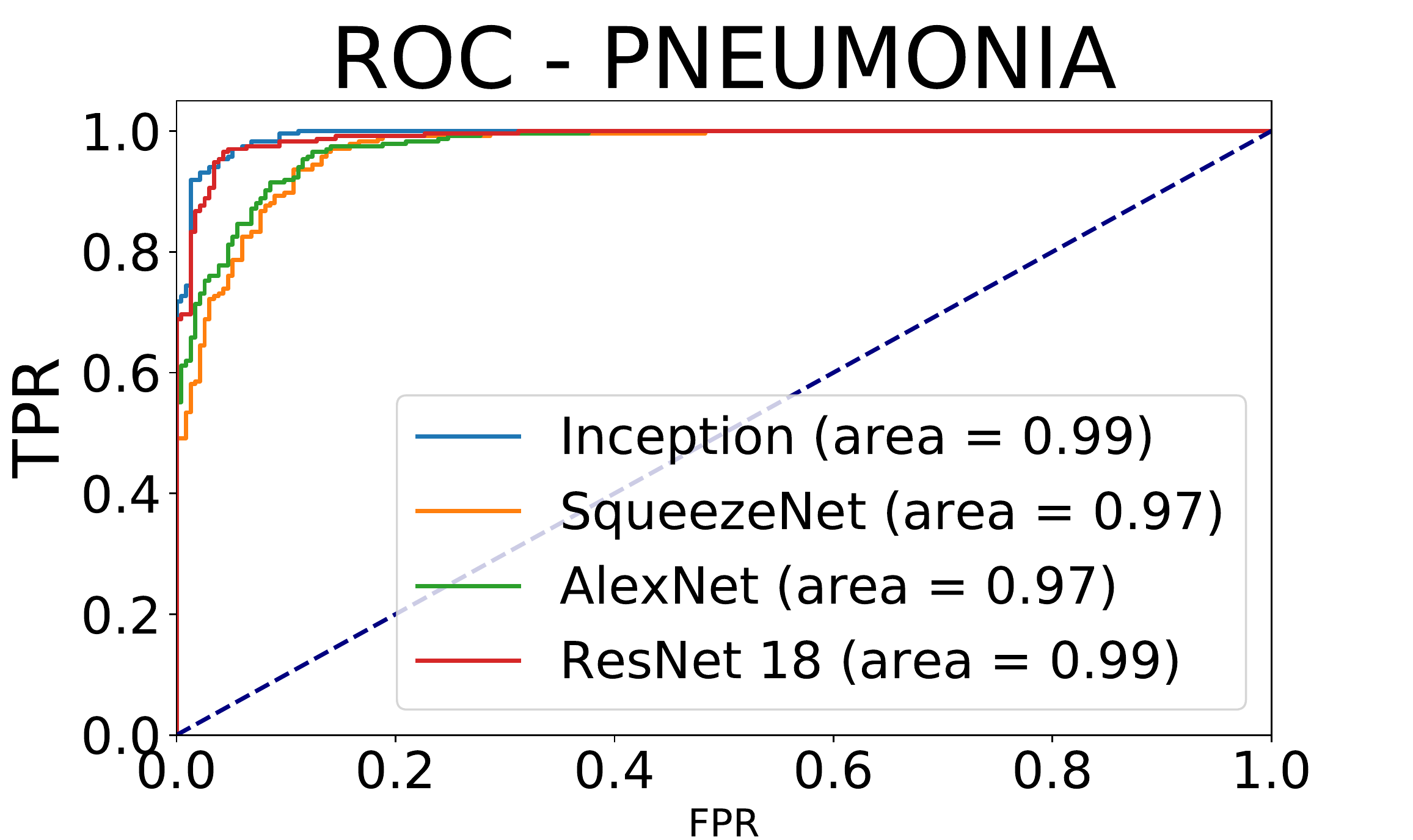}}
\caption{ROC curves for networks}
\label{grafico_roc}
\end{figure}

Fig. \ref{grafico_roc} shows the ROC curves for the networks that explain better the selectivity  of the classifiers. The confusion matrices shows that networks without data augmentation had a better accuracy on the \texttt{Pneumonia} class than the ones with data augmentation but now on the ROC curves we can see that this is not really the case. The data augmentation provided a better separation of the two classes making a more robust classifier reflecting on the ROC curves getting near the northwest corner of the graph.

\section{Conclusions}
It is possible and realistic to create a computer-aided diagnosis (CAD) system using ConvNets even with little computational resources for network training and a small dataset. In the better accuracy cases showed we only needed a few hours to complete the network training and some acceptable results even emerged on the first epochs of training.

To get a reliable CAD system it is convenient to rely on data augmentation techniques to avoid the over fitting problem of ConvNets. In the cases of large datasets maybe this may not be necessary but considering our current dataset of $2,682$ images data augmentation was necessary to get reliable results. One can even argument that the current dataset is not small since in the ImageNet case each class had $1,000$ images and they also used data augmentation techniques to improve the results in \mycite{ImageNet}.
\section*{Acknowledgement}
The author would like to thanks professor Dr. Renato Tinós for his class in Bio-Inspired Computation that helped to grasp the concepts behind neural networks.

\printbibliography

\end{document}